\documentclass[english,runningheads,a4paper]{llncs}
\usepackage{amssymb}
\usepackage{graphicx}
\usepackage{epstopdf}
\usepackage[lined,boxed,commentsnumbered]{algorithm2e}
\usepackage{algpseudocode}
\usepackage[latin9]{inputenc}
\usepackage{listings}
\usepackage{babel}

\usepackage{xcolor}

\usepackage[pdfborder={0 0 0}]{hyperref}

\usepackage{url}
\urldef{\mailsa}\path|{rcanirudha, sddhrthrt, avinashnb141}@gmail.com, sowmyakamath@nitk.ac.in|
\newcommand{\keywords}[1]{\par\addvspace\baselineskip
\noindent\keywordname\enspace\ignorespaces#1}

\renewcommand{\algorithmicrequire}{\textbf{Input:}}
\renewcommand{\algorithmicensure}{\textbf{Output:}}

\begin{document}

\mainmatter  % start of an individual contribution

\title{A Framework for Web Services Retrieval Using Bio Inspired Clustering}

\titlerunning{Web Services Retrieval Using Bio Inspired Clustering}

\author{Anirudha R C \and Siddhartha R Thota \and Avinash N Bukkittu  \and Sowmya Kamath S}
\authorrunning{Web Services Retrieval Using Bio Inspired Clustering}

\institute{Department of Information Technology, National Institute of Technology Karnataka,\\
Mangalore, India, 575025\\
\mailsa\\}

%\toctitle{Lecture Notes in Computer Science}
%\tocauthor{Authors' Instructions}
\maketitle

\begin{abstract}
Efficiently discovering relevant Web services with respect to a specific user query has become a growing challenge owing to the incredible growth in the field of web technologies. In previous works, different clustering models have been used to address these issues. But, most of the traditional clustering techniques are computationally intensive and fail to address all the problems involved. Also, the current standards fail to incorporate the semantic relatedness of Web services during clustering and retrieval resulting in decreased performance. In this paper, we propose a framework for web services retrieval that uses a bottom-up, decentralized and self organising approach to cluster available services. It also provides online, dynamic computation of clusters thus overcoming the drawbacks of traditional clustering methods. We also use the semantic similarity between Web services for the clustering process to enhance the precision and lowerthe recall. 
\end{abstract}

\section{Introduction}
In recent years, there has been a massive growth in the field of web technologies as organizations are increasingly moving towards providing services to their consumers over the Web. This has resulted in an increase in the number of Web services and applications using these services in e-commerce, travel and other domains are also growing exponentially. This in turn creates a huge demand for several techniques for efficiently find the Web services. 

	UDDI Universal Business Registries (Universal Discovery, Description and Integration) which were intended to be a standard for publishing and finding web services has been shut down in 2005 due to lack of popularity. The main aim of the public UDDI initiative was to allow users and applications alike to find Web services through keyword searches. However, the number of Web services retrieved based on purely syntactic keyword searches can be quite large and most of search results were irrelevant to users, resulting in  lower precision and high recall values. This is because keywords are described in natural language which has free form and provides ample scope for ambiguity in terms of meaning and context. Syntactically different words can be semantically similar (synonyms). Likewise, semantically different words can be syntactically similar (homonyms). Keyword based searches do not capture the underlying semantics of Web services. The limitations of syntax based Web service discovery paved the way for research involving semantics for search and retrieval of Web services. A semantic approach is intuitive in the sense that it tries to meaningfully discover relevant Web services.
	
	Various studies have shown clustering to be a more effective way for discovery of Web services. But as the Web services are ever increasing in number, incorporating a static clustering algorithm for their discovery demands the recomputation of all the clusters every time a new service is added. Due to the highly computationally intensive nature of performing semantic comparison between Web services, the use of such techniques is not feasible. This calls for a clustering technique which can easily accomodate newly added Web services without the need for significant recomputation of already formed clusters. Previously such dynamic algorithms have been proposed to handle the problems of large datastreams like network packet flow, clickstream data, etc. This paper presents a framework to incorporate a similar dynamic algorithm for Web service discovery.
	
	In order to address the issues around the discovery and retrieval of Web services, we propose a framework which is based on bio-inspired clustering to cluster available services dynamically to facilitate their efficient retrieval. The proposed model is inspired by the bio-inspired FlockStream Algorithm proposed by Foresterio et al~\cite{FPS09} which proposes a model to compute clusters on the fly without affecting the previously formed clusters. Adapting this technique for clustering Web services addresses the problems of recomputation of all clusters. Also, we include the concept of semantic similarity between the services ensuring better precision and recall during retrieval. Furthermore, we also describe a technique to retrieve the required Web services from these clusters based on the creation of a virtual Web service formed using tags from the requirements given by the user as a query for retrieval. We refer to this as a virtual boid or virtual agent.

%	 In order to solve these problems, we use a bio-inspired clustering algorithm for efficiently clustering and finding Web services on the Web. A clustering framework for Web services has a costly comparison algorithm between services and has services added in real time. To avoid having to compute the clustering all the way on addition of a new service and to enhance the performance of the clustering, we adapt a bio-inspired multi-agent system that is decentralized, bottom-up and self-organizing proposed by Foresterio et al \cite{FPS09}. In implementing the algorithm for web services, we create a search mechanism by creating virtual service elements that represent the required parameters in the query, and then letting that virtual service element be clustered using this algorithm. We obtain the results by using the clusters formed with the virtual service. Since the algorithm is bottom-up and self-organizing, the overhead in search operation is reduced to a great extent.%
	 
	 The rest of the paper is organised as follows: In Section \ref{sec:Related Work} we discuss related work in the domain of clustering web services. Section \ref{sec:Methodology} presents a detailed discussion on the proposed framework, describing the steps involved in the clustering process. Finally, Section \ref{sec:Conclusion} concludes the paper along with scope for future work in the proposed framework.

\section{Related Work} \label{sec:Related Work}

	Much research has been carried out in the area of discovery of Web services. Of them clustering for reducing the search space for relevant service retrieval has gained much popularity in recent years. In this section, we provide a brief overview of existing research in the domain of clustering of Web services. 
	
	Dong et al. \cite{DHM04} proposed a clustering algorithm for semantically cluster the Web services, based a set of similarity search primitives and algorithms for implementing these primitives. However, their technique showed a significant increase in precision and recall.  In the work of Nayak et al. \cite{NL07} Wordnet and OWL-S ontology \cite{wordnet} was utilized to improve the syntactic description provided by WSDL. Further, clustering was used to group the Web services based on these enhanced descriptions.
	
	Abramowicz et al. \cite{AH07} presented an architecture involving elements chained in a workflow for semantic Web services clustering and filtering which complement each other and also help in solving some problems of Service Oriented Architecture paradigm. An extensible directory system having a flexible query language was designed by Constantinescu et al. \cite{CB05} to cluster the Web services into hierarchical categories allowing the efficient matching of constraints to retrieve results incrementally.
Ma et al. \cite{MZ08} used a semantic approach to cluster Web services. Assuming that efficiency can be improved by eliminating irrelevant data, K-means is applied to remove unrelated services. After this, Probabilistic Latent Semantic Analysis (PSLA) was used to exploit semantics behind words describing the Web services. 
	
	Most of the previous works do not address the issue of computational intensivity of the clustering process. Forestiero et al. \cite{FPS09} proposed a clustering method which is density based and efficiently handles data streams by performing an asynchronous and local search. It also involves online cluster management phase which allows users to obtain clusters on demand. This algorithm is based on \emph{flocking model} \cite{KJ01} and though it has not been used in the area of Web services, incorporating the suggested techniques can help address the above issue.
	
	Though various methods have been proposed, they fail to address the issues of handling Web services efficiently as there is no method which is a combination of semantic relatedness and dynamic handling. We now propose one such method which combines the goodness of both in the following section.

\section{Methodology} \label{sec:Methodology}

\begin{figure}[]
			\centering
			\includegraphics[scale=0.51]{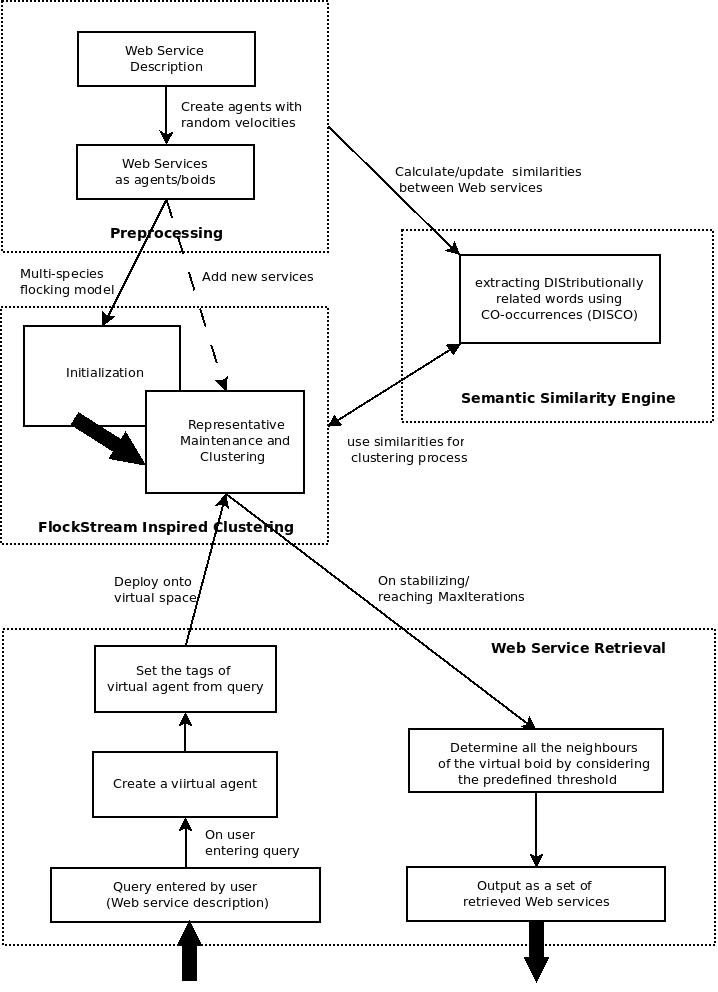}
			\caption{System architecture for proposed framework}
			\label{fig:System diagram for proposed framework}
		\end{figure}

We describe the proposed novel framework for the efficient discovery and retrieval of Web services in this section. The proposed model can be viewed as consisting of three main processes.  Firstly, the \emph{semantic similarity} between all the available Web services is calculated. Secondly, \emph{clustering} is performed on the services by the clustering algorithm, inspired from the flockstream algorithm using the calculated similarities. Finally, \emph{retrieval} of the relevant services from the obtained clusters by entering a query which describes the user requirements. Each of these processes are detailed in the sections below. Figure \ref{fig:System diagram for proposed framework} provides the system architecture of the proposed framework.
		
\subsection{Semantic similarity between Web services} \label{sec:meth1}
Using only syntactic methods for calculating the similarity between Web services may not result in clusters having contextually related Web services. This will result in increased recall and lower precision during the retrieval. A large number of Web services will be retrieved and most of them will be irrelevant to the users thus resulting in reduced recall and precision values. For efficient clustering and thus effective retrieval, we must take into account the semantics involved and hence it is essential to incorporate certain semantic similarity measures. In order to achieve this, we use DISCO \cite{DISCO,Kolb08,Kolb09} to measure the semantic similarity between the Web services. 

DISCO (extracting DIstributionally related words using CO-occurrences) is a Java API which when given two arbitrary words provides their semantic similarity. This similarity is obtained by performing statistical analysis on its large text corpus and set of collocations. Hence, using this to compute the similarity will ensure that the grouping of the services is done by taking into consideration the semantics. This is highly relevant here as different service providers may use varied conventions and words to describe their service. The services are represented by tags which describe them. Firstly we calculate the similarities between the corresponding tags of the services using DISCO. Then semantic similarity between Web services is obtained by summing these individual tag similarities and finally normalising the similarity value. Thus, the semantic similarity between Web services is the sum of semantic similarities between the corresponding tags which describe them.

\subsection{Web services as a multi-agent framework}

We use the bio-inspired flockstream algorithm provided in \cite{FPS09} as the basis for the clustering process in the proposed framework. In a novel approach to clustering web services, we model the web services as a flock of independent agents or boids that interact with each other in virtual space. By applying a heuristic strategy inspired by the behaviour of birds, the agents are then clustered using a bottom up,  decentralized and self-organizing method.

Similar to the flocking model in \cite{CR} we simulate the animation of Web services in virtual space. These agents are steered in the space with the help of three steering velocities which operate parallely on each agent at every instance of time: \emph{separation}(\overrightarrow{v_{sp}}, moving away from colliding neighbour), \emph{alignment}(\overrightarrow{v_{al}}, moving towards the average heading and velocity of the neighbours), \emph{cohesion}(\overrightarrow{v_{ch}}, moving towards the neighbours). These rules also consider only a small locality of every agent, which makes it an efficient local-search based algorithm.

\vspace{3mm}

\hspace{20mm}$\overrightarrow{v_{al}} = \displaystyle\sum_{i}^{n}(\overrightarrow{v_{i}}) $

\vspace{3mm}

\hspace{20mm}$\overrightarrow{v_{sp}} = \displaystyle\sum_{i}^{n}(\frac{\overline{\overrightarrow{v_{i}}+\overrightarrow{v_{c}}}}{d(n_{i},a_{c})}) $

\vspace{3mm}

\hspace{20mm}$\overrightarrow{v_{ch}} = \displaystyle\sum_{i}^{n}(p_{i}-p_{c}) $

\vspace{3mm}

where $\overrightarrow{v_{i}}$ is the velocity of the $i^{th}$ neighbour $n_i$, $\overrightarrow{v_{c}}$ is the velocity of the current agent, $d(n_{i},a_{c})$ is the distance between the current agent $a_{c}$ and the neighbour $n_{i}$, $p_{i}$ and $p_{c}$ are the positions of $n_i$ and $a_c$ respectively.

  Apart from the above velocities, there are two more forces acting on the agents that make use of the similarity measure \emph{Sim}($n_{i}$, $a_{c}$) calculated based on the similarity of the web services using the technique described in the Section \ref{sec:meth1}. The attraction force which keeps two similar agents together:
  
\vspace{3mm}  

\hspace{20mm}$v_{sim}=\displaystyle\sum_{i}^{n}(Sim(n_{i},a_{c})*d(p_{i},p_{c}))$ 

\begin{flushleft}$d(p_{i},p_{c})$ is the Eucledian distance between $p_{i}$ and $p_{c}$ in virtual space.

\vspace{3mm}

Likewise dissimilar agents repel each other with a force:

\end{flushleft}

\vspace{3.5mm}

\hspace{20mm}$v_{dsim}=\displaystyle\sum_{i}^{n}\frac{1}{(Sim(n_{i},a_{c})*d(p_{i},p_{c}))}$ 

\vspace{4mm}

\begin{flushleft}This total velocity $\overrightarrow{v}$ is the velocity of the agent in the virtual space. \end{flushleft}

This model has two phases as described in the Flockstream algorithm, \emph{Initialization} and \emph{Representative Maintenance and Clustering}. Initialization phase is the one in which the Web services are added for the first time. Each service is introduced into the virtual space as a boid or agent with random velocities. The agents in the space then follow the rules of the flocking algorithm \cite{FPS09} to form clusters. Representative Maintenance and Clustering involves performing the clustering for a previously decided number of iterations on introduction of a new set of Web services. This phase generates clusters on the fly without having to recompute all the earlier clusters. In both the phases, the movements of the agents are governed by the steering velocities and forces described earlier. 

\subsection{Web service retrieval}

In addition to the above process, we describe the retrieval process to answer search queries for web service discovery. A query for discovery of Web service is defined as a request from the user that has a desired query string for each of the descriptive property (tags) of the Web service. When such a query is submitted to the framework, a virtual agent representing an imaginary service is created. The tags of this service will be set by filling in the corresponding descriptions from the user query. The virtual agent corresponding to this service is then deployed into the virtual space with a random velocity. The process now continues similar to the Representative Maintenance and Clustering phase of the Flocking algorithm and the virtual agent interacts with its neighbours and finally joins a cluster or the process is stopped after a maximum number of iterations. At this instance distance $d(n_i,b)$ is computed between all the agents in the space $n_i$ with the virtual agent $b$. All the agents having distance $d(n_i,b)$ lesser than $\epsilon$, a predefined constant, are considered as neighbours and are returned as the results of the search query. The pseudo code for the described retrieval process is shown in Algorithm \ref{alg:Pseudo code for search operation for Web service discovery}.   

\vspace{2mm}

		\begin{algorithm}[h]
		
% 			\SetLine % For v3.9
 			\vspace{1mm}
			 \algorithmicrequire{ A query with search parameters} \\ 
			 \vspace{1mm}
			 \algorithmicensure{ A set of Web services that match the query} \\
			  \vspace{3mm}
			  
			 	new Agent virtualAgent \\
			 	 \vspace{1mm}
			 	virtualAgent.properties = query.fields \\
			 	 \vspace{1mm}
				insertNewAgent(virtualAgent)
				 \vspace{1mm}
			 	
 				\For{$i=1 ... MaxIterations$}
 				{
 				\vspace{1mm}
					\If{(neighbours(virtualAgent) $>$ numResults)}
					{
						\vspace{1mm}
   						return neighbours(virtualAgent)
   						\vspace{1mm}
   					}
   					\vspace{1mm}
 				}
 				\vspace{1mm}
 						
 			\label{alg:Pseudo code for search operation for Web service discovery}
 			 			
 			\caption{Pseudo code for Web service retrieval}
 			
		\end{algorithm}

\section{Conclusion and Future Work} \label{sec:Conclusion}

\hspace{5mm}	This paper proposes a bio-inspired approach for clustering Web services based on the \emph{flocking model} \cite{KJ01}. We present the framework  that eliminates the need for offline cluster computation and generates the clusters on the fly. Hence, it is also possible to dynamically add of Web services efficiently by reducing the complexity and time required for the reclustering process. Also, this model uses DISCO for the calculation of similarity as it captures the semantics behind the words describing the services that helps improve the recall and precision. Finally, it also provides a way to retrieve the required services effectively which can be achieved through the introduction of a virtual agent which is dervied from the tags in the user entered query.

	In the algorithm presented, virtual space is considered to be an infinite space. Hence, the time required to obtain stabilized clusters is high which needs to be addressed. Also, currently we need to perform an exhaustive search of the entire virtual space to retrieve the agents which are within the sensor range of the agent under consideration. This is computationally intensive and hence slow. This calls for an optimized and efficient way to represent and retrieve the neighbouring agents. 
	
	Though DISCO allows the calculation of semantic similarity between the web services, it fails to recognise the tags when they are not given in their basic forms. This results in inaccurate calculation of similarity thus affecting the purity of the clusters adversely. Also, abbreviations are usually not found in DISCO's text corpus, and even if they are, the value obtained does not give a true picture of semantic similarity between the words. Hence, there needs to be more work in this area to ensure higher accuracy and more semantic relatedness among the clustered services.
 	
 	Finally, the proposed model may be optimized to work with real world Web services and provide a better method to determine the sensor range and the weights associated with the velocities under consideration.

\bibliographystyle{ieeetr}	
\bibliography{report}
\end{document}